\pdfoutput=1

\documentclass[11pt]{article}

\usepackage[dvipsnames]{xcolor}

\usepackage{ACL2023}

\usepackage{times}
\usepackage{latexsym}

\usepackage[T1]{fontenc}

\usepackage[utf8]{inputenc}

\usepackage{microtype}

\usepackage{inconsolata}

\usepackage{booktabs}
\usepackage{amssymb}
\usepackage{multirow}
\usepackage{comment}
\usepackage{paralist}
\usepackage{graphicx}
\usepackage{colortbl}
\colorlet{tablerowcolor}{gray!5}
\newcommand{\rowcol}{\rowcolor{tablerowcolor}} %
\usepackage[per-mode = fraction]{siunitx}

\title{Natural Language Generation for Advertising: A Survey}

\author{
    Soichiro Murakami, \ Sho Hoshino, \ Peinan Zhang\\
  CyberAgent, Inc., \\
  \texttt{\{murakami\_soichiro,hoshino\_sho,zhang\_peinan\}@cyberagent.co.jp}
}

\begin{document}
\maketitle
\begin{abstract}
Natural language generation methods have emerged as effective tools to help advertisers increase the number of online advertisements they produce.
This survey entails a review of the research trends on this topic over the past decade, from template-based to extractive and abstractive approaches using neural networks.
Additionally, key challenges and directions revealed through the survey, including metric optimization, faithfulness, diversity, multimodality, and the development of benchmark datasets, are discussed.
\end{abstract}

\section{Introduction}
Online advertising\footnote{In this survey, advertisements produced in various media are referred to simply as ``ads'' for convenience.} plays a crucial role in enabling online vendors to promote their products and services, and has seen a rapid growth over the past decade.
To this end, natural language generation for advertising (AdNLG) has emerged to meet the demand for automating ad-creation processes and has been studied mainly by industry-related researchers~\cite{Hughes2019-sh,Kamigaito2021-iy,golobokov2022deepgen}.
However, AdNLG has received considerably limited scholarly attention when compared to other domain-specific areas of NLG, such as dialogue systems for healthcare~\cite{valizadeh-parde-2022-ai}.
This is because AdNLG lacks systematically organized information such as a comprehensive study of task definitions and benchmark datasets, leading to a barrier for new entrants to the field.
As a result, industry misses the opportunity for potentially helpful collaborations with academia, whereas academia overlooks attractive research topics in the ad domain.
This situation is problematic for both academia and industry.


\begin{figure}
    \centering
    \includegraphics[keepaspectratio, width=0.95 \linewidth]{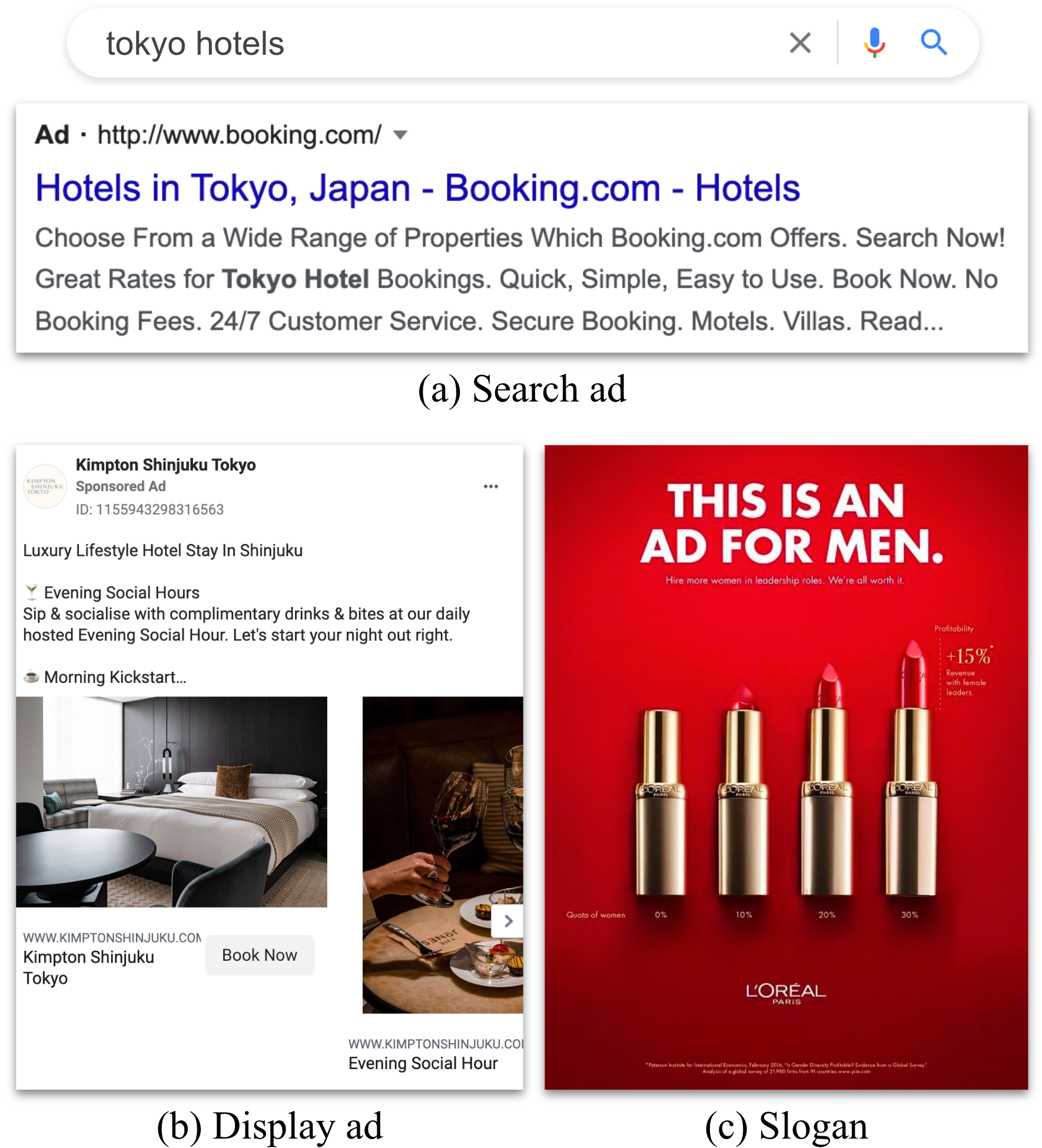}
    \caption{Examples of various online advertising media.}
    \label{fig:ad-example}
\end{figure}

In this study, we investigated recent progress in AdNLG to provide a comprehensive and introductory literature review to solve this problem for both worlds.
Considering standard practices in NLG, we collected papers published primarily at international conferences.
Subsequently, we screened the collected papers according to the AdNLG criteria defined in \S\ref{sec:background} and \S\ref{sec:review-method}.
The literature was divided into the three categories according to the approach used: template-based, extractive, and abstractive approaches.
Five key aspects of AdNLG have been identified in the literature: advertising performance, diversity, faithfulness, fluency, and relevance.
Furthermore, we reviewed the evaluation methods used in offline and online.

Finally, we discuss the challenges and future directions elucidated through this survey for AdNLG, including metric optimization, diversity, faithfulness, multimodality, and the development of benchmark datasets.
We hope that this survey will encourage more academic research in the field of AdNLG and promote the implementation of cutting-edge industrial applications.

\section{Background}\label{sec:background}
The survey covered three types of advertising investigated in the selected studies, namely search engine advertising, display advertising, and slogans (Figure \ref{fig:ad-example}).
The survey introduces their respective characteristics and differences along with some measures referred to as advertising performance metrics as background information.

\subsection{Search Engine Advertising}
Search engine advertising, or search ad, displays ads with or alongside search engine results that are expected to be relevant to user queries (Figure \ref{fig:ad-example}a).
In general, search ads are sold based on the keywords selected by advertisers.
These keywords are also known as \textit{bid words} and are developed as part of advertising campaigns; typically, they are related to advertisers' landing pages (LPs)\footnote{A landing page is a web page that users reach after clicking an ad advertising products and services.} describing the advertised business.
Advertisers also provide ad texts to attract users' attention, which mainly refer to both keywords and LPs.
When user queries match a given keyword, ad delivery platforms such as Google Ads select an ad text through a bidding phrase and then display the resulting ad to the user~\cite{wang2015tutorial}.
The most important characteristic of search ads is that advertisers can directly reach users, as they can display a search ad to a user only once that user has entered a specified matching keyword. 

\subsection{Display Advertising}\label{sec:display-ad}
Display advertising is primarily offered in a banner format in designated advertising spaces on a website or application (Figure \ref{fig:ad-example}b).
Banners can present various types of appeals because a combination of text, images, videos, and other media may be included.
In contrast to search ads, display ads are more suitable for engaging with broader audience with access to websites or applications.

\subsection{Slogans}\label{sec:slogan}
Slogans\footnote{A more detailed definition of various slogans used in AdNLG is provided in Appendix \ref{sec:appendix}.} refer to any phrase used to attract Internet users' interest in products, services, or campaigns (Figure \ref{fig:ad-example}c).
Unlike search and display ads, slogans aim to create long-lasting impressions on target users.

\subsection{Advertising Performance Metrics}\label{sec:ad-performance}
The ultimate goal of AdNLG is to increase the effectiveness of ads, which is measured using advertising performance metrics collected by ad delivery platforms~\cite{zhu2017cpc}, such as click-through rates (CTR) and cost per click (CPC).
CTR measures the average number of direct user feedback interactions; a high CTR value indicates better engagement.
In contrast, CPC measures the average cost charged to deliver a given ad; a low CPC value indicates more potential revenue for advertisers.
The performance of the search and display ads can be measured using CTR or CPC.
However, quantifying slogan performance is challenging.

\section{Literature Review}\label{sec:review-method}
In this section, we briefly explain the literature review process.
The following two sections provide an overview of the approaches and evaluation methods used for AdNLG.

First, we surveyed literature over the past decade by searching for relevant works published primarily at international conferences using queries such as ``\textit{ad text}.''
Next, 24 out of 53 papers were screened according to our criteria, such as the definition of ads in \S\ref{sec:background} and the exclusion of closely related lines of work such as headline generation.
Furthermore, we examined the statistics of affiliation types using selected papers.

\paragraph{Search}
We considered studies published at international academic conferences since 2019 as our primary resource.\footnote
{In particular, conferences hosted by ACL (e.g., ACL, EMNLP, and NAACL), ACM (e.g., CIKM, KDD, and The Web Conf), and other venues such as CVPR were considered.}
Additionally, we gathered preprints, including those from academic journals, to provide a reference for ongoing discussions in related fields.
Using search queries such as ``\textit{ad text}'' and the definition of ads in \S\ref{sec:background}, we initially obtained 53 papers from the search.

\begin{table*}[t]
\footnotesize
\tabcolsep 3pt
\centering
\begin{tabular}{lccccccccc}
    \toprule
    \multicolumn{1}{c}{\multirow{2}[3]{*}{Work}} &
        \multicolumn{3}{c}{Approach} &
        \multicolumn{6}{c}{Key Aspects} \\ \cmidrule(lr){2-4} \cmidrule(lr){5-10} 
    \multicolumn{1}{c}{} &
        \multicolumn{1}{c}{Template} &
        \multicolumn{1}{c}{Extractive} &
        \multicolumn{1}{c}{Abstractive} &
        \multicolumn{1}{c}{Ad Perf.} &
        \multicolumn{1}{c}{Diversity} &
        \multicolumn{1}{c}{Faith.} &
        \multicolumn{1}{c}{Fluency} &
        \multicolumn{1}{c}{Relevance} &
        \multicolumn{1}{c}{Misc.} \\ \midrule
                \citet{chai2022fast}           & -- & -- & \checkmark & -- & \checkmark & -- & \checkmark & -- & -- \\
        \rowcol \citet{golobokov2022deepgen}   & -- & \checkmark & \checkmark & -- & \checkmark & \checkmark & \checkmark & -- & -- \\
                \citet{jin2022slogan}          & -- & -- & \checkmark & -- & \checkmark & \checkmark & \checkmark & -- & -- \\
        \rowcol \citet{kanungo2022cobart}      & -- & -- & \checkmark & \checkmark & -- & -- & \checkmark & -- & -- \\
                \citet{li-etal-2022-culg}      & -- & -- & \checkmark & -- & -- & -- & \checkmark & -- & -- \\
        \rowcol \citet{wei2022creater}         & -- & -- & \checkmark & \checkmark & -- & -- & \checkmark & --  & -- \\
                \citet{alnajjar_toivonen_2021} & \checkmark & -- & -- & -- & -- & -- & -- & -- & \checkmark \\
        \rowcol \citet{duan2021query-variant}  & -- & -- & \checkmark & -- & -- & -- & -- & \checkmark & -- \\
                \citet{Kamigaito2021-iy}       & -- & -- & \checkmark & \checkmark & -- & -- & \checkmark & \checkmark & -- \\
        \rowcol \citet{kanungo-etal-2021-ad}   & -- & -- & \checkmark & -- & -- & -- & \checkmark & --  & -- \\
                \citet{Wang2021-uq}            & -- & -- & \checkmark & \checkmark & -- & \checkmark & \checkmark & --  & -- \\
        \rowcol \citet{zhang2021chase}         & -- & -- & \checkmark & -- & -- & -- & \checkmark & --  & -- \\
                \citet{mishra2020refinement}   & -- & -- & \checkmark & \checkmark & -- & -- & -- & --  & -- \\
        \rowcol \citet{youngmann2020}          & -- & -- & \checkmark & \checkmark & -- & \checkmark & -- & --  & -- \\
                \citet{Hughes2019-sh}          & -- & -- & \checkmark & \checkmark & -- & -- & -- & --  & -- \\
        \rowcol \citet{wang2019quality}        & -- & -- & \checkmark & \checkmark & -- & -- & -- & --  & -- \\
                \citet{thomaidou2013}          & \checkmark & -- & -- & -- & -- & -- & -- & \checkmark &  -- \\
        \rowcol \citet{thomaidou2013grammads}  & -- & \checkmark & -- & -- & -- & -- & -- & --  & \checkmark \\
                \citet{Fujita2010-xm}          & -- & -- & \checkmark & -- & -- & -- & -- & --  & \checkmark \\
        \rowcol \citet{Bartz2008-ke}           & \checkmark & -- & -- & -- & -- & -- & -- & \checkmark  & -- \\
                \bottomrule
    \end{tabular}%
\caption{Three major approaches of AdNLG, along with five key aspects identified in the literature review.
Miscellaneous (Misc.) denotes the studies that do not fit into any of these aspects.}
\label{tab:approaches}
\end{table*}

\paragraph{Screening}
Next, we performed a two-stage screening process, eventually selecting only 24 out of the 53 papers, as follows:
\begin{inparaenum}[(i)]
\item We selected 40 papers in the first stage by filtering out papers on topics unrelated to AdNLG, such as click prediction, keyword generation, and surveys.
\item Next, we closely read and selected the 24 most related papers in the second stage.
We excluded closely related lines of work from our main scope, including headline generation \cite{kanungo-etal-2021-ad}, product description \cite{shao2021vae}, and snippet generation \cite{yi2022effective}, with the exception of a few studies \cite{kanungo-etal-2021-ad,kanungo2022cobart}.
\end{inparaenum}

\paragraph{Affiliation}
An interesting finding is that most studies were from or related to the advertising industry.
Among the 40 papers selected in the first stage, 53\% (21 papers) had at least one industry-affiliated author.
For the 24 papers selected in the second stage, this number increased to 88\%, reflecting that only three studies \cite{thomaidou2013grammads,thomaidou2013,alnajjar_toivonen_2021} were conducted by academic researchers.
This finding will be discussed later in \S\ref{sec:benchmark}.

\section{Approaches}
\label{sec:approaches}
In this section, we provide an overview of the major approaches and current research directions for AdNLG identified in the literature review.
Table \ref{tab:approaches} presents 20 out of the 24 papers selected by the screening process~(\S\ref{sec:review-method}).\footnote{Of the four papers not presented in Table \ref{tab:approaches}, two of them are outside the main scope of our study, as mentioned in \S \ref{sec:review-method}, and the remaining two are discussed in \S \ref{sec:evaluation}.}
In particular, we introduce three major approaches in this field: template-based, extractive, and abstractive approaches.
Additionally, we present a brief introduction to the current research directions based on five key aspects of AdNLG: advertising performance, diversity, faithfulness, fluency, and relevance, as well as their motivations.

\paragraph{Template-based Approach}\label{sec:template}
The template-based approach refers to the traditional methods in NLG \cite{Bartz2008-ke} that fill appropriate words, such as keywords (e.g., \textit{Vehicle}), into slots in a predefined template (e.g., \textit{Purchase Your Dream} \underline{\hspace{3em}}) to generate an ad text (e.g., \textit{Purchase Your Dream \underline{Vehicle}}).
The template-based approach can generate grammatically correct and attractive ad texts because the templates are manually designed by ad creators.
However, constructing numerous such templates creatively is labor-intensive, and this approach does not scale well when ad texts require considerable linguistic variation.
Therefore, several attempts have been made to automate the creation of templates from existing ad texts \cite{alnajjar_toivonen_2021} and extraction of keywords to be filled into slots from LPs~\cite{thomaidou2013}.

\paragraph{Extractive Approach}\label{sec:extractive}
The extractive approach refers to extractive document summarization methods \cite{thomaidou2013grammads}, in which important sentences from the input such as an LP are selected and the sentences are output as ad texts.
The extractive approach can generate ad text that is consistent with the input because it extracts text from input LPs.
However, \citet{golobokov2022deepgen} noted that the extractive approach is more suitable for generating ad titles than ad descriptions.

\paragraph{Abstractive Approach}\label{sec:abstractive}
The abstractive approach refers to generating new and unique ad text that accurately captures the content of a given input.
AdNLG using this approach is often formulated as a text-to-text generation task, including document summarization \cite{Hughes2019-sh,Kamigaito2021-iy} and sentence rewriting \cite{Fujita2010-xm,mishra2020refinement}, inspired by various ad-creation processes.
For instance, \citet{Hughes2019-sh} regarded the task as document summarization and used an LP as the source document. 
In contrast, \citet{Fujita2010-xm} regarded the task as sentence compression and sentence summarization in which an ad text is produced by trimming and reconstructing the dependency tree of a given sentence, respectively.

With the recent success of neural networks in a wide variety of NLG tasks, abstractive approaches using neural language generation models have been more actively studied~\cite{Hughes2019-sh,Kamigaito2021-iy,Wang2021-uq,golobokov2022deepgen} than the other two approaches.
Therefore, we mainly discuss the key aspects of the abstractive approach in the following sections.

\subsection{Advertising Performance}
\label{sec:ctr}
The goal of AdNLG, as described in \S \ref{sec:ad-performance}, is to increase the effectiveness of ads. 
To achieve this goal, writing ad texts that are attractive to the audience is essential to persuade them to click the ad.
Therefore, numerous studies have incorporated advertising performance metrics as an objective function to directly optimize AdNLG methods.

Reinforcement learning (RL) has been widely used as a representative approach to directly optimize advertising performance metrics.
\citet{Hughes2019-sh}, a pioneer in generating ads via RL, employed a CTR prediction model as a reward function with self-critical RL \cite{rennie2017scst} to optimize the predicted CTR (pCTR).
Similarly, \citet{Kamigaito2021-iy} considered the prediction of another metric based on the ad quality score\footnote{\url{https://support.google.com/google-ads/answer/6167118}}.
\citet{kanungo2022cobart} applied self-critical RL to pretrained Transformer models \cite{vaswani2017attention} such as BART \cite{lewis-etal-2020-bart} and \citet{Wang2021-uq} proposed an RL algorithm called a masked sequence policy gradient that integrates efficiently with large pretrained models and effectively explores the action space.

Alternatively, \citet{wei2022creater} employed contrastive learning, which encourages a model to generate ad texts with a higher CTR.
They created pairs of positive and negative ad texts in terms of CTR through an online A/B test and applied contrastive learning to enhance the discriminative ability of the model for ad texts with different CTR.
In addition, \citet{wang2019quality} proposed a quality-sensitive loss function based on CTR to distinguish between training samples of different qualities during model training. 
First, they constructed a CTR-based latent space as a quality indicator to quantitatively measure the quality score of each training sample between the source and target texts.
Next, they used the quality score to reduce the likelihood of generating low-quality text by weighting the cross-entropy objective.

Furthermore, \citet{kanungo2022cobart} proposed a method to optimize ad texts with control codes; they grouped the observed CTRs into multiple categorical buckets that are used as additional control codes for input.
\citet{mishra2020refinement} and \citet{youngmann2020} formulated AdNLG as sentence rewriting, wherein an existing ad text is provided as input and a refined version with a higher CTR is the generated output.
To construct text pairs for this task, \citet{mishra2020refinement} considered two ad texts in the same ad group as an input--output pair (an existing--optimized text pair) if the corresponding CTRs differed by more than 10\%.

\subsection{Diversity} \label{sec:diversity}
In online advertising, a phenomenon called \textit{ad fatigue}~\cite{abrams2007fatigue} negatively affects the effectiveness of ads; this phenomenon corresponds to when users see the same ads frequently and lose interest.
This motivates advertisers to produce diverse ad texts.

Formally, \citet{tevet-berant-2021-evaluating} defined the diversity used in NLG as the spectrum between the following two ends:
\begin{inparaenum}[(a)]
\item \textit{form} diversity, which corresponds to the word surface (how to say it?), and
\item \textit{content} diversity, which corresponds to the meaning of words (what to say?).
\end{inparaenum}
These aspects of diversity are not orthogonal; therefore, changing one aspect may lead to changes in the other.

Regarding form diversity, \citet{jin2022slogan} proposed a model designed to generate slogans using syntactic control.
Their method generates syntactically diverse slogans by appending part-of-speech tags such as nouns (\texttt{NN}) and verbs (\texttt{VB}) to the beginning of an input sequence.

Regarding content diversity, \citet{golobokov2022deepgen} and \citet{chai2022fast} proposed essentially the same controllable ad text generation model that can highlight different selling points from an input LP.
They applied an ad category classifier that predicts the aspects of selling points for an ad text (such as \textit{Product or Service}, \textit{Advertiser Name or Brand}, and \textit{Location}) to their training data and controlled the content of the generated ad texts via control codes in the inference phase. 

\begin{table*}[!t]
\centering
\footnotesize
\begin{tabular}{lcccccc}
\toprule
\multicolumn{1}{c}{\multirow{3}[4]{*}{Work}} & \multicolumn{5}{c}{Offline} & Online \\
\cmidrule(lr){2-6} \cmidrule(lr){7-7}
& Reference-based & \multicolumn{3}{c}{Reference-free} & Human \\
\cmidrule(lr){2-2} \cmidrule(lr){3-5}
& Word Overlap & Diversity & Language Model & Domain-specific \\
\midrule
\citet{chai2022fast} & \checkmark & -- & -- & \checkmark & \checkmark & -- \\
\rowcol \citet{golobokov2022deepgen} & \checkmark & \checkmark & -- & -- & \checkmark & \checkmark \\
\citet{jin2022slogan} & \checkmark & \checkmark & -- & \checkmark & \checkmark & -- \\
\rowcol \citet{kanungo2022cobart} & \checkmark & -- & -- & \checkmark & -- & -- \\
\citet{li-etal-2022-culg} & \checkmark & -- & -- & -- & -- & -- \\
\rowcol \citet{wei2022creater} & \checkmark & -- & -- & -- & \checkmark & \checkmark \\
\citet{alnajjar_toivonen_2021} & -- & -- & -- & -- & \checkmark & -- \\
\rowcol \citet{duan2021query-variant} & \checkmark & \checkmark & -- & \checkmark & \checkmark & -- \\
\citet{Kamigaito2021-iy} & \checkmark & \checkmark & \checkmark & \checkmark & \checkmark & \checkmark \\
\rowcol \citet{kanungo-etal-2021-ad} & \checkmark & -- & \checkmark & -- & \checkmark & -- \\
\citet{Wang2021-uq} & \checkmark & -- & \checkmark & \checkmark & \checkmark & \checkmark \\
\rowcol \citet{zhang2021chase} & -- & \checkmark & \checkmark & \checkmark & \checkmark & \checkmark \\
\citet{mishra2020refinement} & \checkmark & -- & -- & \checkmark & -- & \checkmark \\
\rowcol \citet{youngmann2020} & -- & -- & -- & \checkmark & -- & \checkmark \\
\citet{Hughes2019-sh} & \checkmark & -- & -- & \checkmark & \checkmark & -- \\
\rowcol \citet{wang2019quality} & \checkmark & -- & -- & -- & \checkmark & -- \\
\citet{thomaidou2013grammads} & -- & -- & -- & -- & \checkmark & \checkmark \\
\rowcol\citet{thomaidou2013} & -- & -- & -- & -- & \checkmark & -- \\
\citet{Fujita2010-xm} & -- & -- & -- & \checkmark & \checkmark & -- \\
\rowcol\citet{Bartz2008-ke} & -- & -- & -- & \checkmark & \checkmark & -- \\
\bottomrule
\end{tabular}
\caption{Overview of conventional evaluation criteria for AdNLG.}
\label{tab:eval}
\end{table*}

\subsection{Faithfulness} \label{sec:faithfulness}
Recent studies have reported that generated texts often contain unfaithful information called \textit{hallucinations}\footnote{Following  \citet{10.1145/3571730}, we refer to the often ambiguously used terminology \textit{hallucination} as unfaithfulness to the input, unlike other definitions such as the distinction between \textit{factual} and \textit{non-factual} hallucinations \cite{cao2022hallucinated}.} \cite{wei2022creater,golobokov2022deepgen,Wang2021-uq}.
For example, \citet{wei2022creater} reported that approximately \numrange{5}{10}\% of generated texts contain such hallucinations.
This can be a significant barrier to practical applications because unfaithful information may be presented to users.
Therefore, improving the faithfulness of generated text is key to the widespread implementation of ad text-generation systems in the advertising industry.

To mitigate this problem, recent studies have suggested several approaches by borrowing ideas from hallucination--mitigation methods in document summarization.
For instance, \citet{golobokov2022deepgen} introduced phrase-based data filtering to remove unfaithful ad texts from training data containing content that is not present in an input LP.
They used a list of potentially erroneous phrases and patterns obtained by studying human evaluation for data filtering.
Furthermore, they reported that applying phrase-based filtering to both the training data and generated texts significantly improved the faithfulness of the output texts.
\citet{Wang2021-uq} proposed an RL reward function that constraints unfaithful words or phrases to prevent an RL-based model from generating attractive but unfaithful phrases such as ``{\it free shipping}'' and ``{\it official site}'' \cite{wang2013psychological}.
They observed a significant increase in the percentage of generated text containing unfaithful claims when the proposed RL function was not employed.

Furthermore, several studies used an entity masking technique to alleviate the entity hallucination problem \cite{jin2022slogan,youngmann2020}.
For example, \citet{jin2022slogan} first performed named entity recognition on ad texts and then replaced each entity token with a placeholder such as \texttt{[person]}.
Once the ad text was generated, they performed a dictionary lookup to replace the placeholder tokens with the corresponding original surface forms.

\subsection{Fluency} \label{sec:plm}
Low-quality text, in terms of grammar and fluency, does not properly convey information to users. 
Thus, advertisers must ensure that the ads are of a high quality.

Pretrained language models (PLMs), such as GPT-2 \cite{radford2019language}, have demonstrated remarkable progress in fluency on a wide variety of NLG tasks and are widely used to generate fluent text in AdNLG \cite{jin2022slogan,Wang2021-uq,kanungo-etal-2021-ad,kanungo2022cobart,chai2022fast,golobokov2022deepgen}.
Following the considerable success of document summarization, BART~\cite{lewis-etal-2020-bart} and UniLM~\cite{dong2019unilm} have been widely used in this field.
In addition, some studies have introduced domain-specific PLMs trained on corpora related to an e-commerce domain such as marketing descriptions, user reviews, and advertising texts \cite{zhang2021chase,wei2022creater,li-etal-2022-culg}.

In addition to the aforementioned approaches, \citet{Kamigaito2021-iy} introduced fluency reward into an RL-based model.
They calculated a fluency reward consisting of two scores: a grammatical score calculated using a language model and a length score, which represents the appropriateness of the length of the generated texts. 

\subsection{Relevance}\label{sec:copy}
In AdNLG, the relevance of ad text to user preferences is a crucial factor in determining the effectiveness of ads. 
Relevant ad text is more likely to capture user attention and interest; thus, it results in higher click rates and better overall performance.
Therefore, several studies have attempted to increase the relevance of ad text to user search intentions.

A natural way to improve relevance is to include keywords such as search queries and bidding phrases in the ad text.
Because these keywords are closely related to user search intent, keyword insertion can directly improve the relevance to user preferences. 
In fact, ad delivery platforms such as Google Ads also encourage advertisers to include keywords in ad text to increase the effectiveness of ads.\footnote{\url{https://support.google.com/google-ads/answer/6238826}}
In this context, the template-based approach \cite{Bartz2008-ke,thomaidou2013} is the simplest and most certain way to improve relevance, because it allows arbitrary keywords to be filled into slots.
In contrast, using the abstractive approach, \citet{Kamigaito2021-iy} introduced keyword relevance as a reward function for an RL-based model. 
They designed a reward function for the coverage and positions of keywords in the generated text to initially generate an ad text with important keywords.

Furthermore, to address the many-to-one problem of ad data, where different keywords refer to one general-purpose ad text, \citet{duan2021query-variant} proposed a dynamic association module to expand the receptive field of text-generation models based on external knowledge. 
This module explores and selects suitable associated words related to a user query employing external knowledge.
The associated words serve as a bridge between the high- and low-frequency keywords. 
This enables the model to handle diverse and low-frequency queries more effectively because it helps the model to generalize and transfer from high- to low-frequency keywords.

\section{Evaluation Methods}
\label{sec:evaluation}
In this section, we discuss how to evaluate the automatically generated ad text yielded by the methods reviewed.
Given that NLG evaluation is a challenging topic and discussions on its foundations are ongoing \cite{howcroft-etal-2020-twenty,tevet-berant-2021-evaluating,gehrmann-etal-2022-repairing}, we focus on the evaluation methods practically used in AdNLG, including both general and domain-specific methods, with connections and differences in mind.
We also discuss the lack of publicly available benchmark datasets and two ways of data splitting for AdNLG.

Table \ref{tab:eval} summarizes the evaluation criteria and corresponding literature.
Briefly, two types of evaluation schemes are represented: offline and online.
In terms of NLG, offline evaluations, such as automatic and human subjective evaluations, correspond to intrinsic evaluations, whereas online evaluations correspond to extrinsic evaluations.

\subsection{Automatic Evaluation}
Automatic evaluation methods provide the minimum turnaround time and best reproducibility in exchange for a low correlation with human judgment.
Popular reference-based methods, such as BLEU \cite{papineni-etal-2002-bleu}, ROUGE \cite{lin-2004-rouge}, METEOR \cite{banerjee-lavie-2005-meteor}, and CIDEr \cite{Vedantam_2015_CVPR}, measure the overlap of word $n$-grams between hypotheses and references.

Reference-free methods, including diversity metrics, language models, and domain-specific metrics, have been increasingly applied in this field.
Diversity metrics, such as distinct \cite{li-etal-2016-diversity}, Self-BLEU \cite{10.1145/3209978.3210080}, and Pairwise-BLEU \cite{pmlr-v97-shen19c}, measure the uniqueness of the generated ad text as the number of unique $n$-grams or multiple hypotheses.
Language models are used in two ways: as a reference-free metric of the perplexity \cite{jelinek1977perplexity,brown-etal-1992-estimate} of a given hypothesis, or as a reference-based metric of cosine similarity between the hypothesis and reference.
Domain-specific metrics, such as pCTR, keyword coverage, keyword position, and predicted ad auction win rate \cite{shuai2020prediction}, were used as pseudo-criteria to simulate online evaluation \cite{mishra2020refinement,Kamigaito2021-iy}.

\subsection{Human Evaluation}
Alternatively, human experts can be employed to obtain more accurate and detailed subjective evaluations.
However, this approach requires considerable time and costs.
The human subjective evaluation criteria used in AdNLG include both general NLG criteria \cite{howcroft-etal-2020-twenty}, and domain-specific criteria such as attractiveness.
These criteria can be converted into single scores in two ways:
\begin{inparaenum}[(i)]
\item an evaluation using an absolute scale such as a five-point Likert scale \cite{likert1932}, and by
\item a comparative evaluation that shows hypotheses side-by-side in pairwise, listwise, or more sophisticated ranking systems such as TrueSkill\textsuperscript{\texttrademark} \cite{herbrich2007trueskill}.
\end{inparaenum}
Subjective evaluation criteria used in the literature are provided in Appendix~\ref{sec:criteria}.

\subsection{Online Evaluation}
Instead of relying only on offline evaluation metrics, advertising performance metrics can be directly collected using online evaluation methods, including A/B testing and difference-in-differences \cite{card2000did}.
An ad delivery platform is typically set up and fed automatically generated ad text.
After a certain period, the ad delivery platform yields relevant measures based on the actual user feedback.
Furthermore, causal inference can be performed on factors that may influence user feedback such as seasonal influences.

\subsection{Datasets and Data Splitting}
\label{sec:datasets}
Notably, no conventional benchmarks have been established for AdNLG methods.
This is due to the lack of publicly available datasets with only a few exceptions such as University College London's open advertising dataset\footnote{\url{https://code.google.com/archive/p/open-advertising-dataset/}} \cite{wang2015tutorial} and a slogan dataset\footnote{\url{https://github.com/chaibapchya/slogans}}.
Therefore, most studies shown in Table \ref{tab:eval} were conducted on ad delivery platforms using in-house ad data.

In addition to the lack of benchmark datasets, a few studies \cite{mishra2020refinement,mishra2021tsi,golobokov2022deepgen} reported that conventional data splitting methods such as hold-out validation and cross-validation do not work well for AdNLG.
Their observations suggest that conventional methods cannot mitigate overfitting or potential data leakage such as temporal leakage.

To assess the risk of data leakage, they conducted experiments in two ways.
\begin{inparaenum}[(a)]
\item \textit{Warm-start split}, which is another name for the conventional methods in this context.
\item \textit{Cold-start split}, which is a novel alternative, is tailored purposely for evaluating generalization performance.
Therefore, in terms of generalization, the cold-start is more difficult than the warm-start.
In particular, the dataset was split into two random parts using advertisers as a key index.
The model was trained using a larger portion of the split data.
Subsequently, the remaining smaller portion, which is equal to the unseen advertisers, was fed to evaluate the trained model.
\end{inparaenum}

\section{Discussion and Future Directions}
Following the literature review, we discuss several challenges and future directions found in the literature, including metric optimization, diversity, faithfulness, multimodality, and the development of benchmark datasets.

\subsection{Metric Optimization Needs Evaluation}
Numerous studies \cite{Hughes2019-sh,Kamigaito2021-iy,Wang2021-uq,kanungo-etal-2021-ad,kanungo2022cobart} have utilized reinforcement learning to optimize advertising performance metrics, such as pCTR (\S\ref{sec:ctr}).
In addition, \citet{wang2019quality} proposed a CTR-based loss function and \citet{wei2022creater} used the CTR-driven contrastive learning method.
Therefore, this study attempts to identify how they evaluate the direct optimization of advertising performance metrics.

Surprisingly, the effectiveness of optimizing advertising performance metrics has not been well studied in the literature, with the exception of the ablation study reported by \citet{Kamigaito2021-iy}.
Thus, the conventional evaluation criteria used in AdNLG do not necessarily correspond to the advertising performance metrics.
Another gap in the literature is the lack of error analysis, which refers to a detailed investigation of the components that contribute to a specific improvement in fine-grained analytics.
The issue of standardized error analysis has also been raised by the NLG community in general \cite{howcroft-etal-2020-twenty,gehrmann-etal-2022-repairing}, which is a future direction for AdNLG.

\subsection{Diversity is Twofold}
To overcome the critical ad fatigue~\cite{abrams2007fatigue}, several studies \cite{jin2022slogan,golobokov2022deepgen,chai2022fast} have proposed controllable models that can generate diverse ad texts concerning either the form or content aspect of diversity, as mentioned in \S\ref{sec:diversity}.
Nonetheless, most studies on AdNLG have not considered diversity or distinguished between these two aspects.
For example, only a few studies have incorporated $n$-gram-based diversity metrics into their evaluation criteria (\S\ref{sec:evaluation}), such as distinct and Self-BLEU, which are unsuitable for measuring the content aspect, as suggested by \citet{tevet-berant-2021-evaluating}.
Therefore, fully diversity-aware metrics must be incorporated into the future AdNLG models and evaluation protocols.

\subsection{Measuring Faithfulness is Challenging}
AdNLG models often suffer from hallucinations that are attractive but unfaithful to a given input, such as ``\textit{official site}'' \cite{Wang2021-uq}.
To avoid the delivery of unfaithful ad texts in practical applications, the faithfulness of the generated ad texts must be improved, as mentioned in \S\ref{sec:faithfulness}.

However, existing approaches cannot deal with different wording, such as synonyms and abbreviations, because they rely on the surface matching of words and phrases~\cite{golobokov2022deepgen,Wang2021-uq}.
In particular, a generated text is judged to be unfaithful if certain phrases (such as ``\textit{free shipping}'') in a list of constrained phrases appear in the generated text, but not in the input and reference.
Therefore, more robust approaches to lexical variability, such as an NLI-based method \cite{philippe2022summac}, which defines faithfulness scores as entailment probabilities between the input and generated text, are worth exploring.

\subsection{Input is not Text Information Alone}
Although LPs used as input for AdNLG are known to contain various modalities, such as images, videos, and tables, most studies have addressed AdNLG as a text-to-text generation task, as mentioned in \S\ref{sec:approaches}.
Most existing studies do not use text contained in images or layout information.

Multimodal information can be used as a clue for identifying important or attractive content in LPs.
For example, \citet{golobokov2022deepgen}, one of the few studies on AdNLG, used the visual features of the text to extract ad text candidates from LPs.
Regardless, the attention paid to the topic of multimodality in the field is insufficient, compared to other NLG tasks, such as document summarization \cite{zhu-etal-2018-msmo} and question answering \cite{Mathew_2021_WACV}.
We believe that introducing multimodal information may help models understand LPs better and generate more informative and attractive ad texts.

\subsection{Lack of Benchmark Datasets}\label{sec:benchmark}
As discussed in \S\ref{sec:datasets}, publicly available benchmark datasets for AdNLG are lacking in the relevant literature.
This limits researchers from reporting their progress in an open scientific manner.
Consequently, approximately 90\% of the studies were reported by industry-related researchers (see \S\ref{sec:review-method}).
A more serious consequence is that it repels newcomers from seeking an accessible dataset.
This state of the field is problematic for two reasons.
\begin{asparaitem}
\item \textbf{Replicability.} No direct method for comparing the results of previous studies has been reported.
Therefore, the actual progress in the field is relatively difficult to quantify.
\item \textbf{Noisy Real-World Data.} As in-house ad data collected in the real-world are often noisy and heavily structured; considerable efforts have been made in terms of data cleaning and splitting.
\end{asparaitem}

Therefore, benchmark datasets should be developed to accelerate future research.

\section{Conclusion}
This survey introduced research trends in AdNLG, including abstractive, extractive, and template-based approaches.
Furthermore, we presented an overview of the evaluation methods for AdNLG.
Finally, the key challenges and directions for future research elucidated by the survey were discussed.

\section*{Limitations}
This survey covers only limited types of ads as defined in \S\ref{sec:background}.


\bibliography{custom}
\bibliographystyle{acl_natbib}

\clearpage
\appendix

\onecolumn
\begin{table}[!t]
\centering
\footnotesize
\begin{tabular}{ll}
\toprule
\multicolumn{1}{c}{Work} & \multicolumn{1}{c}{Subjective Evaluation Criteria} \\
\midrule
\citet{Bartz2008-ke} & Ideal \\
\rowcol \citet{Fujita2010-xm} & Appropriateness \\
\citet{thomaidou2013grammads} & Attractiveness, Informativeness, Relevance \\
\rowcol \citet{thomaidou2013} & Attractiveness, Clarity, Relevance \\
\citet{Hughes2019-sh} & Grammar, Human-likeness, Repetition \\
\rowcol \citet{wang2019quality} & Diversity, Interestingness, Relevance \\
\citet{alnajjar_toivonen_2021} & Attractiveness, Grammar, Metaphoricity, Relevance \\
\rowcol \citet{duan2021query-variant} & Attractiveness, Fluency, Informativeness \\
\citet{Kamigaito2021-iy} & Attractiveness, Fluency, Relevance \\
\rowcol \citet{kanungo-etal-2021-ad} & Attractiveness, Correctness, Grammar, Relevance \\
\citet{Wang2021-uq} & Accuracy, Attractiveness, Fluency, Grammar, Human-likeness, Relevance \\
\rowcol \citet{zhang2021chase} & Coherence, Fluency, Grammar, Informativeness \\
\citet{chai2022fast} & Faithfulness, Fluency, Grammar, Human-likeness, Relevance \\
\rowcol \citet{golobokov2022deepgen} & Faithfulness, Grammar, Human-likeness, Relevance \\
\citet{jin2022slogan} & Attractiveness, Coherence, Fluency, Grammar, Relevance \\
\rowcol \citet{wei2022creater} & Faithfulness, Grammar, Informativeness, Relevance \\
\bottomrule
\end{tabular}
\caption{List of human subjective evaluation criteria; the names are normalized following \citet{howcroft-etal-2020-twenty}.}
\label{tab:eval-human}
\end{table}

\section{Terminology of Various Slogans}\label{sec:appendix}
In this survey, a variety of short advertising phrases, including brand slogans, short advertising copies, catchphrases, and taglines, are referred to as slogans for the sake of simplicity.
\begin{asparaenum}[(a)]
\item In the context of the advertising industry, ``slogans'' refers to mottos or phrases that serve as a professional mark for a given brand.
Slogans are often patented and are considered important intellectual property.
\item Advertising copy or ``ad copy'' refers to any text written or composed to be displayed as an ad or within an ad.
\item By contrast, ``catchphrases'' are generally more personal and informal than slogans, are often used in media such as movies or novels and may be linked to specific characters or intellectual properties.
\item ``Taglines'' refers to memorable phrases used to make the target users recall a specific product or a range of products.
\end{asparaenum}

\section{Subjective Evaluation Criteria}
\label{sec:criteria}
Table \ref{tab:eval-human} summarizes the subjective evaluation criteria used in the relevant literature, including both NLG criteria in general \cite{howcroft-etal-2020-twenty} and domain-specific criteria such as attractiveness.
Because these criteria have ambiguities in their naming and meaning, as suggested by \citet{howcroft-etal-2020-twenty}, we normalized the names according to the categories provided in the prior study.

\end{document}